# Diffusion Transformer Model With Compact Prior for Low-dose PET Reconstruction

Bin Huang, Xubiao Liu, Lei Fang, Qiegen Liu, *Senior Member, IEEE,* Bingxuan Li

***Abstract*—Positron emission tomography (PET) is an advanced medical imaging technique that plays a crucial role in non-invasive clinical diagnosis. However, while reducing radiation exposure through low-dose PET scans is beneficial for patient safety, it often results in insufficient statistical data. This scarcity of data poses significant challenges for accurately reconstructing high-quality images, which are essential for reliable diagnostic outcomes. In this research, we propose a diffusion transformer model (DTM) guided by joint compact prior (JCP) to enhance the reconstruction quality of low-dose PET imaging. In light of current research findings, we present a pioneering PET reconstruction model that integrates diffusion and transformer models for joint optimization. This model combines the powerful distribution mapping abilities of diffusion models with the capacity of transformers to capture long-range dependencies, offering significant advantages for low-dose PET reconstruction. Additionally, the incorporation of the lesion refining block and penalized weighted least squares (PWLS) enhance the recovery capability of lesion regions and preserves detail information, solving blurring problems in lesion areas and texture details of most deep learning frameworks. Experimental results demonstrate the effectiveness of DTM in enhancing image quality and preserving critical clinical information for low-dose PET scans. Our approach not only reduces radiation exposure risks but also provides a more reliable PET imaging tool for early disease detection and patient management.***

## I. Introduction

Positron emission tomography (PET) is a highly sensitive and widely utilized functional imaging modality employed in the diagnosis, staging, and treatment monitoring of various diseases, including malignant tumors, neurological disorders, and cardiovascular conditions [1-3]. This imaging technique allows for the visualization and quantification of metabolic processes in the body, providing critical information that is essential for accurate medical decision-making [4]. Despite its invaluable role in clinical practice, PET imaging has the inherent drawback of exposing patients to ionizing radiation. This exposure poses significant health risks, particularly for vulnerable populations such as pediatric patients and individuals requiring frequent diagnostic procedures [5-7]. Consequently, reducing the radiation dose in PET imaging is a crucial objective to minimize these risks [8]. However, reducing the dose typically results in increased noise and compromised image quality, creating significant challenges for accurate clinical interpretation [9]. The elevated noise levels in low-dose PET images can obscure critical diagnostic details, leading to potential misdiagnosis or the need for repeat scans, which defeats the purpose of dose reduction [10]. Therefore, there is a pressing need to develop advanced image reconstruction techniques that can enhance the quality of low-dose PET images while maintaining diagnostic accuracy [11].

While traditional denoising methods like Gaussian denoising [12], total variation (TV) [13-14], and non-local means (NLM) [15-16] have been utilized for low-dose PET imaging, undoubtedly, recent advancements in deep learning techniques have shown impressive performance. In recent years, the advent of deep learning has revolutionized the field of medical imaging, offering new avenues for improving image quality of PET reconstruction. Deep learning approaches [17-18], particularly convolutional neural networks (CNNs), have demonstrated remarkable success in various image processing tasks due to their ability to learn complex patterns and features from large datasets. For instance, Gong *et al.* [19] developed a deep learning-based approach to enhance the quality of low-dose PET images by training a CNN to map low-dose images to their full-dose counterparts, significantly reducing noise and improving image clarity. Similarly, Xu *et al.* [20] proposed an encoder-decoder residual CNN framework with concatenate skip connections that yielded satisfactory results at significantly low dose. Other deep learning methods have also shown promise in PET image reconstruction. For example, Peng *et al.* [21] applied a deep learning-based denoising method to low-dose PET images, achieving substantial noise reduction while preserving important image details by incorporating CT information. Gong *et al.* [22] introduced a deep neural network for PET image denoising by using simulation data and fine-tune the last few layers of the network using real data sets. Furthermore, Zhou *et al.* [23] utilized a cycle-consistent GAN (Cycle GAN) to translate low-dose PET images to high-dose equivalents, yielding impressive results in terms of image fidelity and diagnostic quality. Pan *et al.* [24] have demonstrated notable efficacy in the reconstruction of low-dose PET utilizing the contemporary diffusion model.

With the advancement of deep learning, both diffusion models and transformer models have demonstrated significant promise in the field of PET low-dose image reconstruction. The

This work was supported by National Natural Science Foundation of China (62122033, 62201193). (B. Huang and X. Liu are co-first authors.) (Corresponding authors: Q. Liu and B. Li.)

B. Huang is with School of Mathematics and Computer Sciences, Nanchang University, Nanchang, China (huangbin@email.ncu.edu.cn).

X. Liu and Q. Liu are with School of Information Engineering, Nanchang University, Nanchang, China ({liuxubiao, liuqiegen}@ncu.edu.cn).

L. Fang is with Department of Biomedical Engineering, Huazhong University of Science and Technology, Wuhan, China (lei.fang@digital-pet.com).

B. Li is with Institute of Artificial Intelligence, Hefei Comprehensive National Science Center, Hefei, China (libingxuan@iai.ustc.edu.cn).

objective of PET low-dose imaging is to reduce the radiation exposure to patients while maintaining image quality, which poses substantial challenges. Deep learning techniques, particularly diffusion models and transformer models, have shown effectiveness in addressing these challenges by enhancing image quality and preserving essential diagnostic details. Diffusion models, which are generative models that learn the underlying data distribution by simulating the diffusion process, have recently gained attention for their ability to denoise and enhance PET images. Gong *et al.* [25] demonstrated the potential of using denoising diffusion probabilistic models to reduce noise in PET images, showing that employing MR prior as the network input while embedding PET image as a data-consistency constraint can achieve better performance. Similarly, Jiang *et al.* [26] proposed an unsupervised PET enhancement approach based on latent diffusion models, which effectively enhanced PET images with latent information. Another notable contribution by Xie *et al.* [27] involved a dose-aware diffusion model tailored for 3D low-dose PET denoising, validated through multi-institutional studies and real low-dose data. Transformer models, on the other hand, have been widely recognized for their ability to capture long-range dependencies and process large-scale data efficiently. In the context of medical imaging, transformer models have been leveraged to improve PET image reconstruction by exploiting their powerful feature extraction and representation capabilities. Zhang *et al.* [28] introduced the spatial adaptive and transformer fusion network (STFNet) for low-count PET blind denoising with MRI. Additionally, Luo *et al.* [29] developed a 3D transformer-GAN for high-quality PET reconstruction. Hu and Liu [30] presented TransEM, a residual Swin-transformer-based regularized PET image reconstruction method, which further validated the advantages of transformers in PET imaging.

Given the complementary strengths of diffusion and transformer models, there is a growing interest in integrating these approaches to leverage their respective advantages. By combining the iterative refinement capabilities of diffusion models with the rapid processing power of transformer models, it is possible to achieve enhanced reconstruction outcomes for PET low-dose imaging.

In this study, we have proposed a novel diffusion transformer model for low-dose PET reconstruction, called DTM. Specifically, horizontal and vertical compact priors, called joint compact prior (JCP), are extracted from normal-dose PET. This JCP will guide the training process of DTM. DTM comprises both transformer and diffusion stages. During reconstruction procedure, the lesion refining block is seamlessly integrated with penalized weighted least squares (PWLS) to heighten fidelity. This block will identify lesion locations by selecting high-value regions. By leveraging data consistency stage (DCS) to govern the generation process of normal-dose PET, the lesion refining block not only bolsters model interpretability but also ensures the preservation of authenticity and reliability, thereby further enhancing its overall performance.

The theoretical and practical contributions of this work are summarized as follows:

- Integration of diffusion and transformer stages for low-dose PET reconstruction significantly enhances the model's ability to generate normal-dose PET. While the diffusion model enhances the detail reconstruction capability, the transformer model introduces multi-head attention mechanism and global information which makes DTM achieves remarkable reconstruction results.
- Implementation of a lesion refining block will perform targeted reconstruction specifically tailored to these lesions within DCS. By integrating this block with PWLS algorithm, we significantly enhance the recovery capability of lesion regions, achieving positive strides in lesion recovery.
- Introduction of the JCP extraction block allows simultaneous extraction of horizontal and vertical compact priors from normal-dose PET. The diffusion process only needs to predict JCP instead of the entire normal-dose data. This JCP refers to latent information that significantly reduces computational complexity and shortens the reconstruction time required.

The rest of the manuscript is organized as follows. Relevant background on score-based diffusion model is demonstrated in Section II. Detailed procedure and algorithm of the proposed method are presented in Section III. Experimental results and specifications about the implementation and experiments are given in Section IV. At last, conclusion is drawn in Section V.

## II. PRELIMINARY

### A. Vision Transformers

The transformer model, initially developed for natural language processing tasks, has been adapted for various vision tasks including image recognition, segmentation, and object detection. Vision transformers [31] decompose images into sequences of patches, enabling them to learn relationships between patches effectively. These models exhibit a notable ability to capture long-range dependencies between sequences of image patches and adapt to varying input content [32]. Consequently, transformer models have been explored for low-level vision tasks such as super-resolution, image colorization, denoising, and deraining. However, the self-attention mechanism in transformers can lead to quadratic increases in computational complexity with the number of image patches, limiting its application to high-resolution images. To address this challenge, recent methods for low-level image processing adopt alternative strategies to mitigate complexity. One approach involves employing self-attention within local image regions [33-34] using the Swin transformer design [34]. However, this approach restricts context aggregation to local neighborhoods, deviating from the primary motivation behind utilizing self-attention over convolutions and rendering it less suitable for image restoration tasks.

### B. Diffusion Models

Diffusion models have emerged as frontrunners in both density estimation [35] and sample quality enhancement [36]. These models utilize parameterized Markov chains to optimize the lower variational bound on the likelihood function, enabling them to generate target distributions with greater accuracy compared to alternative generative models.

The diffusion process operates on an input image $x_0$, gradually transforming it into Gaussian noise $x_t \sim \mathcal{N}(0, I)$

through $t$ iterations. Each iteration of this process is described as follows:

$$q(x_t|x_{t-1}) = \mathcal{N}(x_t; \sqrt{1-\beta_t}x_{t-1}, \beta_t I) \quad (1)$$

where $x_t$ denotes the noised image at time-step $t$, $\beta_t$ represents the predefined scale factor, and $\mathcal{N}$ represents the Gaussian distribution. During the reverse process, diffusion models sample a Gaussian random noise map $x_t$, then progressively denoise $x_t$ until it achieves a high-quality output $x_0$:

$$p(x_{t-1}|x_t, x_0) = \mathcal{N}(x_{t-1}; \mu_t(x_t, x_0), \sigma_t^2 I) \quad (2)$$

To train a denoising network $\epsilon_\theta(x_t, t)$, given a clean image $x_0$, diffusion models randomly sample a time step $t$ and a noise $\epsilon \sim \mathcal{N}(0, I)$ to generate noisy images $x_t$ according to Eq. (2).

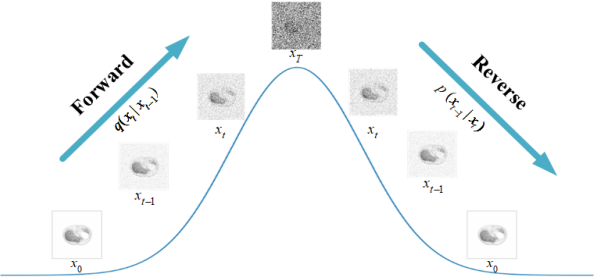

**Fig. 1.** Forward and reverse processes of DDPM.

## III. PROPOSED METHOD

Diffusion model and transformer model each offer unique advantages: diffusion models can fully and efficiently use the powerful distribution mapping abilities to generate images, while transformer models can model long-range pixel dependencies. To effectively combine these two models, it is necessary to make appropriate modifications. Traditional diffusion models require extensive iterative processes, resulting in long training and execution times. Conversely, transformer U-net deliver results rapidly. To merge these models, we can alter the training target of the diffusion model from complete image data to compact priors. This adjustment substantially reduces the training and iteration time for diffusion models, thereby enabling the integration of the two models, as demonstrated in the study by [37].

Inspired by DiffIR [37], we propose a JCP tailored for PET imaging. The JCP serves as a guiding mechanism for the transformer, leveraging the strengths of both the diffusion model and the transformer model to achieve superior reconstruction outcomes. A common challenge in many deep learning models is that noise removal mechanisms may inadvertently lead to the blurring of detailed information and lesion areas. To effectively mitigate this issue, we integrate a lesion refining block with PWLS algorithm. This integration significantly enhances the recovery capability of lesion regions and preserves detailed information, representing a positive stride towards addressing this challenge.

### A. DTM

In summary, DTM consists of four parts: JCP extraction, prior prediction via diffusion process, feeding prior and PET images into transformer stage, and fidelity enhancement using lesion refining block and PWLS block. Initially, a compact prior extraction block is proposed to extract JCP, composed of horizontal prior and vertical prior, from normal-dose PET to guide subsequent low-dose PET reconstruction. This block comprises stacked residual blocks and linear layers.

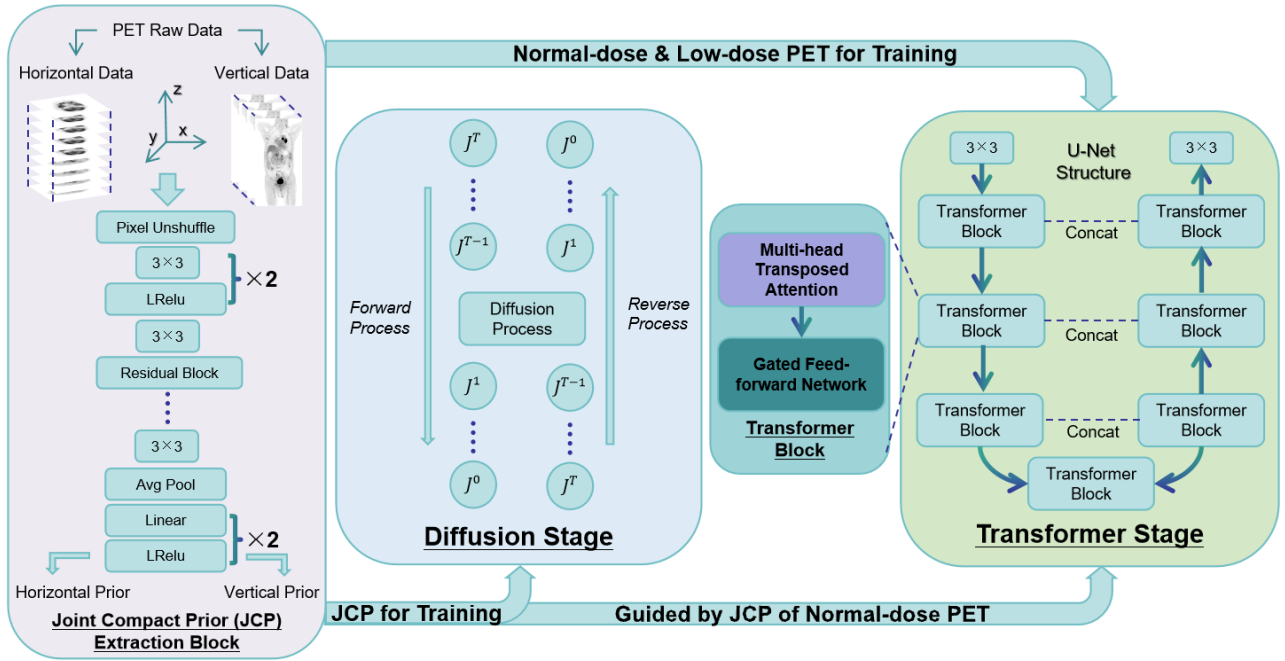

**Fig. 2.** The pipeline of DTM training procedure. DTM mainly consists by JCP extraction block, diffusion stage and transformer stage. The horizontal prior and vertical prior are combined to a JCP which will be feed into the diffusion stage to predict and guide transformer stage to reconstruct final result.

In the diffusion stage, denoised horizontal and vertical compact priors are predicted by the diffusion process. The diffusion module predicts specific noise through conditional diffusion. The estimated degraded prior is then inputted to the transformer module as affine transformation parameters to reconstruct high-quality normal-dose PET.

In the transformer stage, multiple transformer blocks are combined in a U-net form, with each block incorporating a multi-head transposed attention and a gated feed-forward network. The transformer module captures distant pixel interactions, where the multi-head transposed attention module aggregates local and non-local pixel interactions, indicating its ability to perform feature interaction across channels. The gated feed-forward network suppresses less informative features, allowing only useful information to further pass through the network hierarchy. Finally, transformer exploits JCP to restore normal-dose PET from low-dose PET.

### B. Training Procedure

The pipeline of DTM training procedure is shown in Fig. 2. This procedure involves JCP extraction block, diffusion stage and the transformer stage. Initially, the normal-dose PET and low-dose PET are concatenated and downsampled using the PixelUnshuffle operation to serve as the input for the JCP extraction block. Subsequently, JCP is then extracted by the JCP extraction block, denoted as $J$. This JCP is then utilized as dynamic modulation parameters in the multi-head transposed attention and gated feed-forward network of the transformer stage to guide PET training:

$$A' = W_l^1 J \odot \mathrm{Norm}(A) + W_l^2 J \tag{3}$$

where $\odot$ indicates element-wise multiplication, Norm denotes layer normalization, $W_l$ represents linear layer, $A$ and $A'$ are input and output feature maps respectively.

In the multi-head transposed attention, global spatial information is aggregated by projecting $A'$ into query $Q = W_d^Q W_c^Q A'$, key $K = W_d^K W_c^K A'$, and value $V = W_d^V W_c^V A'$ matrices, followed by reshaping and dot-product operations to generate a transposed-attention map. This map is further processed using learnable scaling parameter $\gamma$ and channel separation to generate attention map:

$$\hat{A} = W_c \hat{V} \cdot \mathrm{Softmax}(\hat{K} \cdot \hat{Q}/\gamma) + A \tag{4}$$

where Softmax is an activation function that converts a vector of values into a probability distribution, where each value's probability is proportional to the exponential of the input value, typically used in the output layer of a classification neural network [18]. The gated feed-forward network aggregates local features by employing 1×1 Conv layers to aggregate information from different channels and 3×3 depth-wise Conv layers to aggregate information from spatially neighboring pixels. Additionally, the gating mechanism is applied to enhance information encoding. The gated feed-forward network is characterized by the following process:

$$\hat{A} = \mathrm{GELU}(W_d^1 W_c^1 A') \odot W_d^2 W_c^2 A' + A \tag{5}$$

where GELU, Gaussian error linear unit, is an activation function that uses a smooth, non-linear transformation based on the Gaussian cumulative distribution function to activate neurons in a neural network [38]. The JCP extraction block and transformer stage are jointly trained, enabling the transformer stage to effectively utilize JCP for PET reconstruction.

In the diffusion stage, JCP is trained using the strong data estimation ability of the diffusion stage. The JCP extraction block is utilized to capture $J$, which is then subjected to the diffusion process to sample $J_T$:

$$q(J_T|J) = \mathcal{N}(J_T; \sqrt{\bar{\alpha}_T} J, (1-\bar{\alpha}_T)x) \tag{6}$$

$$\alpha_T = 1 - \beta_t; \bar{\alpha}_T = \Pi_{i=0}^{t} \alpha_i \tag{7}$$

where $T$ is the total number of iterations, $\beta_t$ indicates the predefined scale factor.

In summary, DTM has two training stages: the transformer stage and the diffusion stage. In the transformer stage, DTM directly uses the JCP of the normal-dose PET to guide the training of the transformer, focusing on the complete PET data. In the diffusion stage, only the JCP is trained separately without the need to train the normal-dose PET data.

### C. Reconstruction Procedure

During the reconstruction procedure in Fig. 3, a lesion based DCS is proposed to reconstruct low-dose PET images. It is composed of a lesion refining block combined with the PWLS algorithm. Meanwhile, the JCP-based diffusion stage and transformer stage are also utilized in this procedure.

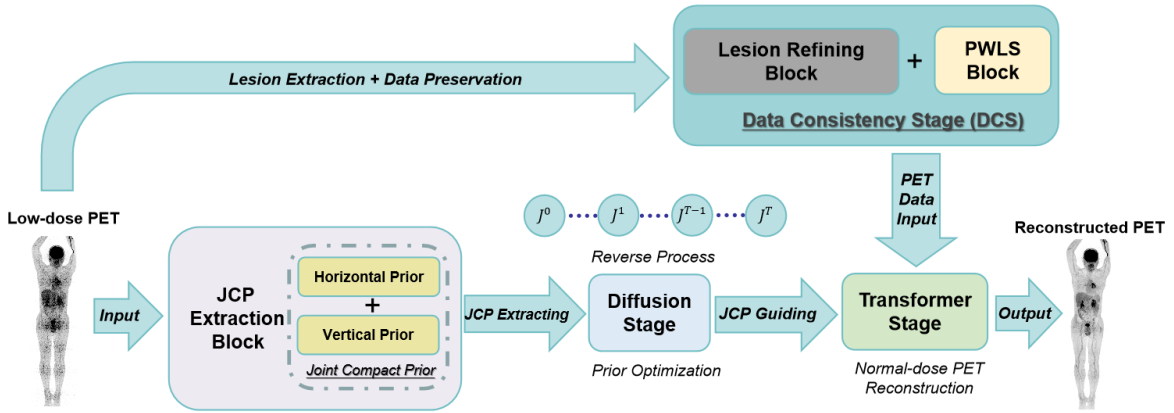

**Fig. 3.** The pipeline of DTM reconstruction procedure. A data consistency stage is involved to inference procedure for lesion refining and detail preserving. JCP of low-dose PET will be reconstructed to JCP of normal-dose PET through diffusion stage. Normal-dose PET will be reconstructed from low-dose PET through JCP guiding and detail preserving by DCS.

DTM initiates from the $T$-th time step and conducts all denoising iterations to derive $\hat{J}$, which is subsequently forwarded to the transformer stage for joint optimization:

$$\hat{J}_{t-1} = \frac{1}{\sqrt{\alpha_t}}(\hat{J}_t - \epsilon \frac{1-\alpha_t}{\sqrt{1-\bar{\alpha}_t}}) \tag{8}$$

where symbol $\epsilon$ represents the same noise, for which we employ the JCP extraction block and denoising process to make predictions. Notably, in contrast to traditional diffusion models, our DTM eliminates variance estimation, which is beneficial for accurate JCP estimation and improved performance.

During the reverse process of the diffusion stage, we initially utilize the JCP extraction block to derive a conditional vector $J$ from the low-dose PET. Subsequently, the denoising process is utilized to estimate noise at each time step, denoted as $\epsilon_t$. This estimated noise is then incorporated into Eq. (8) to obtain $\hat{J}_{t-1}$, which serves as the starting point for the subsequent iteration. Following $T$ iterations, we arrive at the final estimated JCP $\hat{J}$. Subsequently, the transformer module leverages the JCP for the reconstruction of low-dose PET data.

In DCS, we introduce an innovative lesion refining block that automatically extracts lesions from low-dose PET data $x^0$ by identifying high-value regions, which corresponds to areas where the tracer accumulates. Subsequently, target reconstruction $x^i$ is generated, specifically tailored to these lesion refining data $u^i$. By integrating lesion refining block with the PWLS algorithm, we significantly enhance the recovery capability of lesion regions, achieving a positive stride for lesion recovery. The overall optimization equation of DTM can be expressed as follows:

$$x^{i+1} = \underset{x}{\operatorname{argmin}}[\|y - \zeta(u^i)\|_w^2 + \mu R(x^i, \hat{J})]$$
$$s.t. \quad M \odot x^i = v^i$$
$$u^i = [(x^i + \eta v^i)/2] \quad (9)$$

where low-dose PET data $x^0$ is used as input for the reconstruction process to approximate normal-dose data $y$. The operator $\zeta$ is applied for reconstructing the normal-dose PET and $i$ refers to the PWLS iteration number. A weighting $w = diag(1/\sigma^2)$ matrix is constructed based on the standard deviation. The regularization term $R(x^i, \hat{J})$ incorporates JCP information to constrain the solution. Moreover, a high-value region $v^i$ is extracted for refining lesion information, and a binary mask $M$ is used to identify this region, with elements set to 1 for the region and 0 elsewhere. $\mu$ and $\eta$ are modulating factors to maintain the consistency of DTM. The value of $x^{i+1}$ is determined by $x^i$, $u^i$ and $v^i$ during the iterative solving process.

The augmented Lagrangian function can be defined as follows:

$$L(x, u, \lambda_1, \lambda_2) = \|y - \zeta(u^i)\|_w^2 + \lambda_1(M \odot x^i - v^i)$$
$$+ \lambda_2\left(u^i - \frac{x^i + \eta v^i}{2}\right) + \frac{\rho}{2}\|M \odot x^i - v^i\|^2$$
$$+ \frac{\gamma}{2}\|u^i - \frac{x^i + \eta v^i}{2}\|^2 + \mu R(x^i, \hat{J}) \quad (10)$$

where $\lambda_1$ and $\lambda_2$ are Lagrange multipliers. $\rho$ and $\gamma$ are penalty parameters used to control the strength of the penalty imposed by the constraint conditions. Eq. (10) can be broken into two subproblems by updating the variables iteratively:

1) *Subproblem 1* ← Update $x^{i+1}$ by minimizing:

$$x^{i+1} = \underset{x}{\operatorname{argmin}}[\lambda_1(M \odot x^i - v^i) + \lambda_2\left(u^i - \frac{x^i + \eta v^i}{2}\right)$$
$$+ \frac{\rho}{2}\|M \odot x^i - v^i\|^2 + \frac{\gamma}{2}\|u^i - \frac{x^i + \eta v^i}{2}\|^2 + \mu R(x^i, \hat{J})] \quad (11)$$

2) *Subproblem 2* ← Update $u^{i+1}$ by minimizing:

$$u^{i+1} = \underset{u}{\operatorname{argmin}}\ [\|y - \zeta(u^i)\|_w^2 + \lambda_2\left(u^i - \frac{x^i + \eta v^i}{2}\right)$$
$$+ \frac{\gamma}{2}\|u^i - \frac{x^i + \eta v^i}{2}\|^2] \quad (12)$$

Algorithm 1 offers a detailed description of low-dose PET reconstruction, where $\delta$ represents the learning rate, and $\kappa$ represents the convergence threshold.

---

**Algorithm 1: DTM for Reconstruction**

**Require:** $x^0, v^0, u^0, i, w, \zeta, \hat{J}, M, \eta$
1: **Initialization:** $x^0, v^0$, and $u^0$
2: **For** $i = 0$ to 2 do
3:     Update $x^{i+1}$ via Eq. (11)
4:     Update $u^{i+1}$ via Eq. (12)
5:     Take a gradient descent step on $\nabla_x L$
       $x^{i+1} \leftarrow x^i - \delta \nabla_x L$
6:     **If** $\|x^{i+1} - x^i\| < \kappa$
7:     **Else** $i \leftarrow i + 1$
8:     **End if**
9: **End for**
10: **Return** $x^{i+1}$

---

### D. Distinctive Characteristics

We introduce a JCP to guide PET reconstruction through DTM. In the diffusion stage, the compact structure of JCP enables DTM to achieve robust estimations with fewer iterations and smaller model size compared to traditional diffusion models. Traditional diffusion models incur substantial computational costs during iterations, necessitating the random sampling of time steps for denoising optimization. Moreover, the lack of joint training between the denoising process and the decoder means that minor estimation errors from the denoising process can hinder the transformer's full potential. As represented in Eq. (9), JCP is incorporated into the regularization term, enhancing the reconstruction procedure.

Beyond JCP, we also introduce the innovative lesion refining block. This block systematically evaluates all data points from the input PET data, identifying areas that closely resemble lesions. These identified regions are reintroduced into DTM for reconstruction. The reconstructed results are then merged back into the transformer stage using weighting mechanisms to achieve an optimal PET reconstruction outcome. This method addresses a common challenge in many deep learning models, where noise removal may inadvertently lead to the loss of lesion areas. The lesion refining block effectively mitigates this issue by specifically targeting and preserving lesion information during the denoising process, ensuring more accurate and comprehensive recovery of relevant anatomical features in PET images. The functions of lesion refining block can be expressed as the constraint conditions, detailed in Eq. (9). By introducing these constraint conditions, the search space for solutions is minimized, ensuring an optimal solution within the specific constraints. The penalty terms $\frac{\rho}{2}\|M \odot x^i - v^i\|^2$ and $\frac{\gamma}{2}\|u^i - (x^i + \eta v^i)/2\|^2$ can make the optimization problem more proximate to the actual optimal solution.

In summary, our distinctive characteristics include the integration of JCP for guided PET reconstruction and the lesion refining block with PWLS algorithm for preserving critical lesion and detailed structure information. These advancements significantly improve the robustness and accuracy of low-dose PET imaging, offering a more reliable tool for clinical diagnosis and early disease detection.

## IV. Experiments

In this section, we introduce the implementation details of the proposed DTM, as well as the datasets we used for evaluation. Subsequently, the reconstruction results are reported and analyzed. Both quantitative and qualitative evaluations are comprehensively conducted to investigate the performance of DTM. Patient data is used for training and ablation studies, while phantom data is employed for generalization experiments.

### A. Data Specification

The experiment data is generated by the DigitMI 930 PET/CT scanner. This scanner is developed by RAYSOLUTION Healthcare Co., Ltd, and incorporate state-of-the-art all-digital PET detectors. The PET scanner has an axial field-of-view (AFOV) of 30.6 cm within an 81 cm ring diameter.

*Patient data:* A comprehensive data set is utilized, consisting of 19 patients. Each patient undergoes a scan ranging from 4 to 8 beds, with a complete sampling scan time of 45 seconds to 3 minutes per bed. The low-dose PET is performed through resampling at regular intervals. Specifically, the listmode data is segmented based on the chronological sequence, with each cycle representing a 2 millisecond (ms) interval. During each cycle, data within a 1 ms interval is retained, while the remaining data is discarded. Finally, the low-count data is rearranged into image domain. The training data comprises a total of horizontal 25986 2D slices with dimensions of 256×256 and vertical 45384 2D slices with dimensions of 256×144 normal-dose PET data from 12 patients, while the test data is obtained from half dose, quarter dose, and one-tenth dose PET voxels from 7 patients. This study is approved by the institutional review board of the Beijing Friendship Hospital, Capital Medical University, Beijing, China. The approval number is 2022-P2-314-01.

*Phantom data:* To assess image quality using body phantom data, a NEMA body phantom with an interior length of 180 millimeter (mm) was utilized. This phantom contained six fillable spheres with internal diameters of 10 mm, 13 mm, 17 mm, 22 mm, 28 mm, and 37 mm. It was filled with 18F-FDG, having a background activity concentration of 5.3 kBq/ml. The activity concentration in the fillable spheres was four times that of the background.

### B. Model Training and Parameter Selection

In the experiments, DTM is implemented in Python and PyTorch on a personal workstation with a GPU card (NVIDIA RTX-3090-24GB). The proposed approach employs a 4-level encoder-decoder structure. Within the transformer stage, the multi-head transposed attention mechanism utilizes attention heads with configurations of [1, 2, 4, 8], accompanied by respective channel numbers of [48, 96, 192, 384]. Specifically, across levels 1 to 4, the number of dynamic transformer blocks is configured as [3, 5, 6, 6]. Additionally, the number of channels for the JCP extraction block is set to 64. In the diffusion stage, $T$ is set to 4. Adam optimizer is set to $\beta_1 = 0.9$ and $\beta_2 = 0.99$. For PWLS algorithm, $i$ is set to 2. The open-source code is available at: https://github.com/yqx7150/DTM.

### C. Quantitative Indices

To evaluate the quality of the reconstructed data, peak signal-to-noise ratio (PSNR), structural similarity index (SSIM), normalized root mean squared error (NRMSE), contrast ratio (CR) and coefficient of variation (COV) are used for quantitative assessment.

PSNR describes the maximum possible power of the signal in relation to the noise corrupting power. Higher PSNR means better image quality. Denoting $x$ and $y$ to be the estimated reconstruction and the reference image, PSNR is expressed as:

$$PSNR(x, y) = 20\log_{10}[Max(y)/\|x - y\|_2] \quad (13)$$

SSIM is used to measure the similarity between the ground-truth and reconstruction, and it is defined as:

$$SSIM(x, y) = \frac{(2\mu_x\mu_y + c_1)(2\sigma_{xy} + c_2)}{(\mu_x^2 + \mu_y^2 + c_1)(\sigma_x^2 + \sigma_y^2 + c_2)} \quad (14)$$

where $\mu_x$ and $\sigma_x^2$ are the average and variances of $x$. $\sigma_{xy}$ is the covariance of $x$ and $y$. $c_1$ and $c_2$ are used to maintain a stable constant. NRMSE is employed to evaluate the errors and it is defined as:

$$NRMSE(x, y) = \sqrt{\sum_{i=1}^{W} \|x_i - y_i\|_2/W} / (y_{\max} - y_{\min}) \quad (15)$$

where $W$ is the number of pixels within the reconstruction result. If NRMSE approaches to zero, the reconstructed image is closer to the reference image.

The calculation of CR involves comparing the maximum pixel value $M_{lesion}$ within the lesion region of the patient to the mean pixel value $\mu_{liver}$ in the liver region:

$$CR = M_{lesion}/\mu_{liver} \quad (16)$$

COV quantifies the variation in signal intensity across a selected region, like liver, of the reconstructed data. It is computed as the ratio of the standard deviation $\sigma_{liver}$ to the mean $\mu_{liver}$ of the pixel values, providing insights into the uniformity of the reconstructed image:

$$COV = \sigma_{liver}/\mu_{liver} \quad (17)$$

By incorporating CR and COV alongside PSNR, SSIM, and NRMSE, a more comprehensive evaluation of the reconstructed data's quality is achieved, encompassing aspects of contrast, noise, and uniformity.

### D. Experimental Comparison

The proposed DTM is compared with four baseline models for low-dose PET reconstruction including Denoise [14], NLM [13], NCSN++ [39] and Pix2pix-3D [40]. NCSN++ is an unsupervised Score-based diffusion model to generate clear images while Pix2pix-3D is a supervised 3D-GAN to denoise low-dose medical images. The involved parameters are set by the guidelines in their original papers.

In this section, half dose, quarter dose, and one-tenth dose PET images are utilized as input during the inference phase, with normal-dose PET images serving as the ground truth. The PSNR, SSIM and MSE values of the reconstructed results from DigitMI 930 PET/CT scanner are listed in Table I. The best PSNR SSIM and NRMSE values of the reconstructed images are highlighted in bold. In general, DTM presents more details and less noise compared to the other methods. It can be observed that DTM is able to achieve impressive average NRMSE value 0.83 of half dose for 7 patients from Table I. The reconstructed images contain fewer artifacts and more details with decreased noise. Excitingly, the image reconstructed by DTM can reach an average of 47.81 dB in the case of one-tenth dose for 7 patients. In terms of SSIM, the results of DTM are still superior. Thus, DTM can achieve visible gains in terms of noise and artifacts suppression.

TABLE I
RECONSTRUCTION PSNR/SSIM/NRMSE FROM DIGITMI 930 PET/CT SCANNER USING DIFFERENT METHODS AT 50%, 25% AND 10% DOSE.

| PET Dose | Denoise | NLM | NCSN++ | Pix2pix-3D | DTM |
|---|---|---|---|---|---|
| 50% | 55.18/0.8232/1.133 | 55.84/0.8598/1.067 | 55.90/0.8757/1.044 | 56.44/0.8982/0.907 | **56.70/0.9011/0.830** |
| 25% | 51.12/0.8194/1.487 | 50.94/0.8354/1.763 | 51.02/0.8117/1.647 | 52.07/0.8698/1.187 | **52.54/0.8819/1.067** |
| 10% | 45.54/0.7879/2.276 | 45.17/0.8210/2.994 | 45.68/0.7846/2.490 | 46.75/0.8457/1.666 | **47.81/0.8743/1.349** |

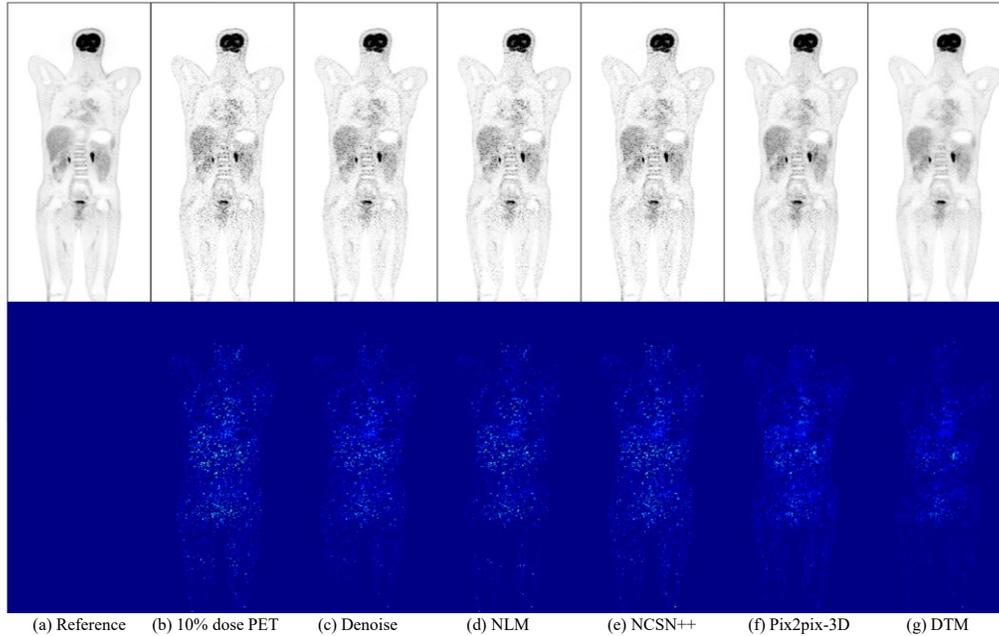

(a) Reference  (b) 10% dose PET  (c) Denoise  (d) NLM  (e) NCSN++  (f) Pix2pix-3D  (g) DTM

Fig. 4. Reconstruction results for coronal images using different methods. (a) The normal-dose PET and (b) 10% dose image versus the images reconstructed by (c) Denoise, (d) NLM, (e) NCSN++, (f) Pix2pix-3D, and (g) DTM. The second row depicts the residuals between the reference and reconstructed images.

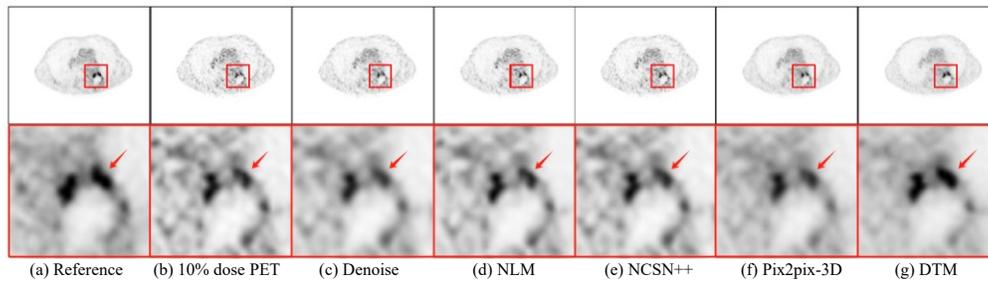

(a) Reference  (b) 10% dose PET  (c) Denoise  (d) NLM  (e) NCSN++  (f) Pix2pix-3D  (g) DTM

Fig. 5. Reconstruction results for transverse images using different methods. (a) The normal-dose PET and (b) 10% dose image versus the images reconstructed by (c) Denoise, (d) NLM, (e) NCSN++, (f) Pix2pix-3D, and (g) DTM. The second row depicts the magnified detail images.

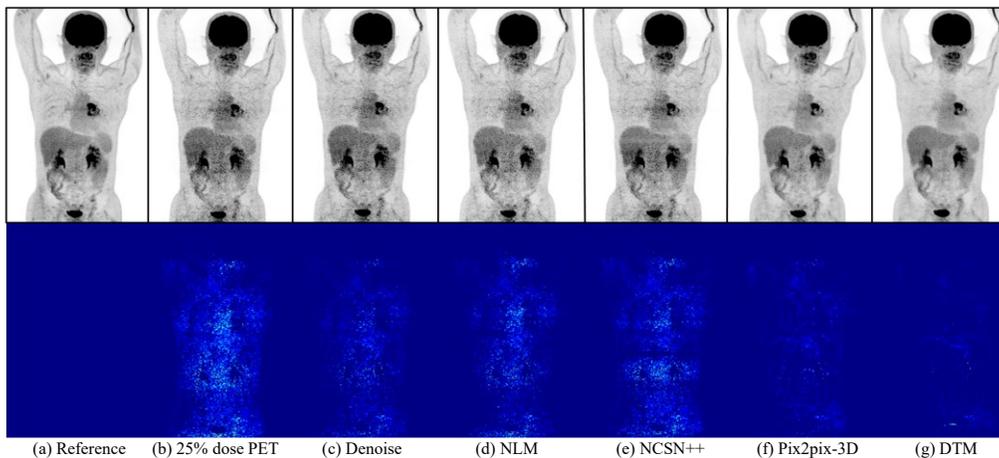

(a) Reference  (b) 25% dose PET  (c) Denoise  (d) NLM  (e) NCSN++  (f) Pix2pix-3D  (g) DTM

Fig. 6. Reconstruction results for MIP images using different methods. (a) The normal-dose PET and (b) 25% dose image versus the images reconstructed by (c) Denoise, (d) NLM, (e) NCSN++, (f) Pix2pix-3D, and (g) DTM. The second row depicts the residuals between the reference and reconstructed images.

To further illustrate the merits of DTM, the reconstructed images and residual images at 10% dose are depicted in Figs. 4-5. The Fig. 5 displays magnified images concerning the transverse graphs, while Fig. 4 presents residual maps for the coronal graphs. As depicted in Figs. 4-5, the Denoise method produces the poorest results due to the excessive blurring of lesions during the denoising process. While Pix2pix-3D showcases commendable denoising capabilities, its ability to reconstruct lesions is inadequate. In contrast, images reconstructed by DTM closely approximate the ground truth, displaying preserved lesions and minimal noise levels.

In the context of coronal maximum intensity projection (MIP) images at 25% dose are depicted in Fig. 6. Since NCSN++ is a 2D unsupervised model, it produces horizontal streaks when generating 3D MIP images, consequently resulting in its poorest performance among the compared algorithms.

Conversely, images reconstructed by NLM and Denoise methods exhibit limited denoising capabilities. Increasing denoising parameters would lead to over-smoothing of images, consequently reducing lesion contrast. Additionally, the Pix2pix-3D method performs moderately well in discerning certain details, ranking second in effectiveness. In contrast, images reconstructed by DTM excel in preserving a broader array of structural details while effectively mitigating streaking artifacts, thereby demonstrating superior performance. These results are consistent with the values presented in Table I.

TABLE II
RECONSTRUCTION CR/COV FROM DIGITMI 930 PET/CT SCANNER USING DIFFERENT METHODS AT 25% DOSE.

| Lesion ID | Normal-dose PET | Low-dose PET | Denoise | NLM | NCSN++ | Pix2pix-3D | DTM |
|---|---|---|---|---|---|---|---|
| 1 | 2.221/0.196 | 2.026/0.294 | 2.013/0.243 | 2.036/0.255 | 2.006/0.224 | 1.924/0.181 | **2.366**/**0.168** |
| 2 | 2.745/0.293 | 2.727/0.542 | 2.210/0.449 | 2.288/0.280 | 2.569/0.407 | 2.465/0.307 | **2.578**/**0.280** |
| 3 | 3.095/0.236 | 4.155/0.483 | 3.656/0.395 | 4.006/0.442 | 3.974/0.271 | 3.359/0.246 | **3.318**/**0.211** |
| 4 | 3.342/0.224 | 3.382/0.463 | 3.044/0.379 | 3.575/0.425 | 3.643/0.255 | 2.4545/0.229 | **3.509**/**0.197** |

Moreover, CR evaluation is performed on 3D regions within five lesion sites across different patients in Table II. The calculation of CR involves comparing the maximum pixel value within the lesion region of the patient to the mean pixel value in the liver region. Remarkably, the CR value of DTM is the closest to that of normal-dose PET compared to the other four methods. A CR value closer to that of normal-dose PET indicates better contrast recovery for lesions. The COV values are calculated in the liver regions. A smaller COV value indicates lower dispersion of image pixel values, meaning that the grayscale or color values are more spatially consistent. This is particularly important in medical image processing and quality control, as a lower COV typically signifies higher image quality and lower noise level.

The profiles generated by Denoise, NLM, NCSN++, and Pix2pix-3D perform adequately at the first peak value but fail to recover the second peak. In contrast, DTM closely matches the ground truth at both peaks, demonstrating the strong profile-preserving capability of the diffusion model. In comprehensive comparison, DTM generates the most accurate profile.

### E. Ablation Study

Through the ablation study, a better understanding of the impact of each component on DTM's performance, as well as their role within the entire model, can be attained. JCP and DCS within DTM will be analyzed. JCP includes vertical and horizontal compact priors, while DCS includes the lesion refining block and PWLS block.

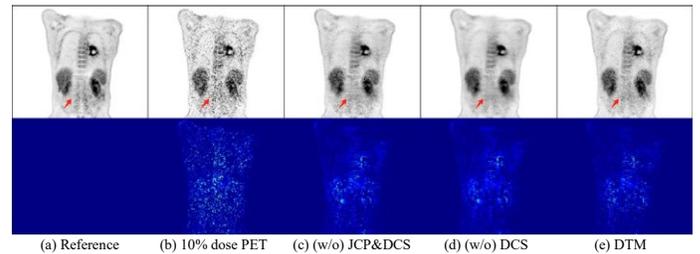
(a) Reference　　(b) 10% dose PET　　(c) (w/o) JCP&DCS　　(d) (w/o) DCS　　(e) DTM
**Fig. 8.** Reconstruction results for coronal images using different methods. (a) The normal-dose PET and (b) 10% dose image versus the images reconstructed by (c) (w/o) JCP&DCS, (d) (w/o) DCS and, (e) DTM. The second row depicts the residuals between the reference and reconstructed images.

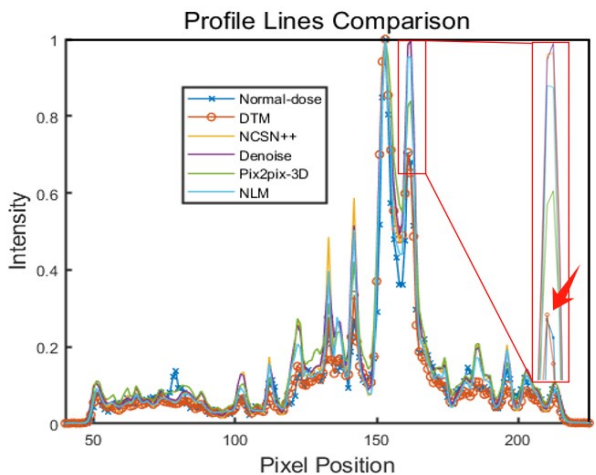
**Fig. 7.** Comparison of PET image profiles between the reference and other four methods on lesion area at 25% dose.

For evaluating the edge-preserving performances, the profile lines for the reconstructed results of different methods are compared in this section. In Fig. 7, a profile line was drawn through the lesion area of a patient, revealing two peak values.

TABLE III
RECONSTRUCTION CR/COV USING DIFFERENT METHODS AT 25% DOSE.

| Lesion ID | (w/o) JCP&DCS | (w/o) DCS | DTM | Normal-dose PET |
|---|---|---|---|---|
| 1 | 1.969/0.182 | 1.898/**0.136** | **2.366**/0.168 | 2.221/0.196 |
| 2 | 2.054/0.236 | 1.924/**0.184** | **2.578**/0.28 | 2.745/0.293 |
| 3 | 3.126/0.124 | 2.499/**0.087** | **3.318**/0.211 | 3.095/0.236 |
| 4 | 2.166/0.112 | 1.899/**0.075** | **3.509**/0.197 | 3.342/0.224 |

Based on Fig. 8 and Table III, it can be observed that JCP

further enhances the model's denoising capability, achieving the lowest COV values. However, this improvement comes at the cost of over-smoothing, which can degrade important image details. In contrast, adding DCS may result in a slight increase in noise but significantly enhances lesion recovery capability, achieving the highest CR values. At last, simultaneously applying JCP and DCS can achieve a balance between noise reduction and detail preservation. This combination results in better lesion recovery and lower noise level.

### F. Generalization and Robustness Analysis

**Phantom Reconstruction Results:** To further validate the robustness of our proposed learning scheme, we change the test data to phantom data and evaluate its generalization performance. The results obtained on the phantom data are presented in Table IV. Impressively, DTM achieves the highest quantitative indices when compared with other comparison methods. In the experiments assessing generalization, the model was not retrained, and the original training model was utilized. This may explain the decline in reconstruction performance of Pix2pix-3D and NCSN++, particularly for Pix2pix-3D. On the other hand, Denoise and NLM demonstrate relatively stable results, with slight fluctuations in the quantitative indices. Conversely, traditional algorithms such as Denoise and NLM can process various datasets because they do not require training. However, the reconstructed result of DTM is 40.49 dB higher than Denoise and NLM respectively. Our DTM outperforms Denoise, NLM, NCSN++, and Pix2pix-3D with higher performance and stronger generalization capability.

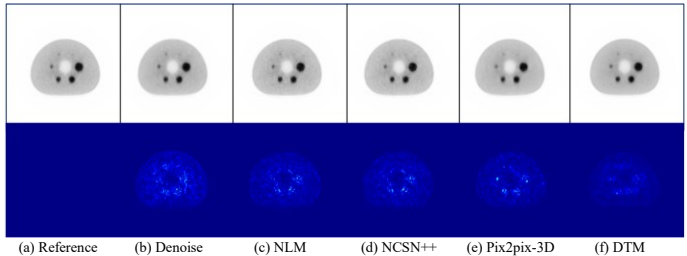

(a) Reference　(b) Denoise　(c) NLM　(d) NCSN++　(e) Pix2pix-3D　(f) DTM

**Fig. 9.** Reconstruction results for phantom data at 50% dose using different methods. (a) The normal-dose PET versus the images reconstructed by (b) Denoise, (c) NLM, (d) NCSN++, (e) Pix2pix-3D, and (f) DTM. The second row depicts the residuals between the reference and reconstructed images.

TABLE IV
RECONSTRUCTION PSNR/SSIM/NRMSE FOR PHANTOM DATA AT 50% DOSE.

| Methods | PSNR | SSIM | NRMSE |
| --- | --- | --- | --- |
| Denoise | 35.28 | 0.7461 | 0.133 |
| NLM | 35.04 | 0.6882 | 0.136 |
| NCSN++ | 32.56 | 0.756 | 0.131 |
| Pix2pix-3D | 34.79 | 0.7158 | 0.138 |
| DTM | **40.49** | **0.8116** | **0.065** |

In Fig. 9, the visual effects of DTM are also exceptional. Pix2pix-3D demonstrates some ability to suppress noises but loses the ability to reconstruct radiated areas. For NCSN++, subtle artifacts and blurred internal structures still persist. In comparison to NCSN++ and NLM, the result of the Denoise shows a slight inferiority. In contrast, DTM approach demonstrates excellent reconstruction capability and effectively compensates for the artifacts in low-dose PET using phantom data. When compared to other competitive reconstruction methods, DTM excels in terms of noise-artifact reduction and detail preservation, offering the best visual results

## V. DISCUSSION

Proposed DTM demonstrates outstanding performance and versatility in various PET imaging tasks. This section will present and analyze the results of normal-dose PET denoising and ultra-low-dose PET reconstruction

**Normal-dose PET Denoising:** Our DTM not only has the capability to restore or reconstruct low-dose PET into normal-dose PET, but also aids in processing normal-dose PET denoising. In Fig. 10, we display zoomed mediastinal position marked by red rectangles. In comparison to normal-dose PET, the outcomes produced by DTM exhibit greater clarity with lower COV calculated in liver area.

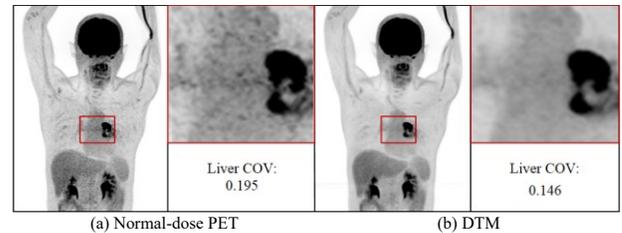

(a) Normal-dose PET　　　　　　(b) DTM

**Fig. 10.** Reconstruction results for MIP images using DTM at normal-dose PET. (a) The normal-dose PET versus the image reconstructed by (b) DTM.

**Ultra-low-dose PET Reconstruction:** When it comes to ultra-low-dose PET reconstruction, DTM can still achieve remarkable results. Without retraining using ultra-low-dose PET, we use DTM to reconstruct from 1% dose PET data. The reconstruction results in Fig. 11 exhibit minimal noise and closely approximate the normal-dose PET, demonstrating the DTM's robust reconstruction capabilities and strong generalization ability.

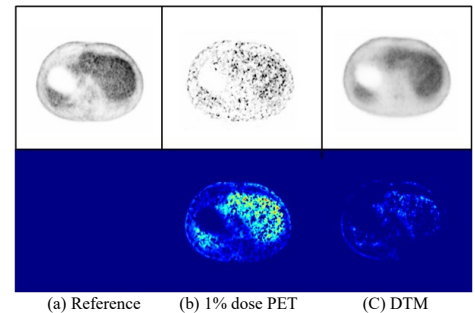

(a) Reference　　(b) 1% dose PET　　(C) DTM

**Fig. 11.** Reconstruction results for transverse images using DTM at 1%-dose PET. (a) The normal-dose PET and (b) 1% dose image versus the image reconstructed by (c) DTM. The second row depicts the residuals between the reference and reconstructed images.

The conducted experiments demonstrate that DTM can serve as a multitask model, capable of denoising normal-dose PET and reconstructing low-dose and ultra-low-dose PET. This showcases its exceptional task adaptability and reconstruction capabilities. The model's versatility allows it to perform excellently across various PET imaging tasks at different dose levels. With further improvements and exploration of DTM, we anticipate uncovering additional applications and potential in PET imaging.

## VI. CONCLUSIONS

Although deep learning-based PET reconstruction methods have achieved significant success in recent years, ensuring high predictive accuracy and robustness of the trained networks remained a challenging issue. In this study, we introduced a novel diffusion transformer approach for low-dose PET reconstruction. Our approach leveraged JCP extraction block to capture the joint prior distribution of normal-dose PET data. Lesion refining block and PWLS were utilized together to preserve the lesion information and detailed structures. Experimental results underscored the effectiveness of the DTM in suppressing the well-known streaking artifacts and preserving crucial image details. These findings positioned DTM as a robust and reliable solution for the challenges associated with low-dose PET reconstruction. Moreover, DTM can serve as a multitask model, capable of denoising normal-dose PET and reconstructing low-dose and ultra-low-dose PET. This study demonstrated that combining the diffusion and transformer models can enhance reconstruction performance and detail richness. Future research endeavors can explore the application of deep learning in real-time PET image reconstruction.